\documentclass[10pt,twocolumn,letterpaper]{article}

\usepackage[pagenumbers]{cvpr} 

\definecolor{cvprblue}{rgb}{0.21,0.49,0.74}
\usepackage[pagebackref,breaklinks,colorlinks,allcolors=cvprblue]{hyperref}

\def\paperID{17209} 
\def\confName{CVPR}
\def\confYear{2026}

\title{
   \vspace{-1.0em} 
   \rule{\linewidth}{3pt} \\ [0.5ex] 
   \textbf{Accelerating Inference of Masked Image Generators via Reinforcement Learning} \\ [0.5ex] 
   \rule{\linewidth}{1pt} 
}

\author{
   \textbf{Pranav Subbaraman$^*$, Shufan Li$^*$, Siyan Zhao, Aditya Grover} \\[0.1in]
   UCLA \\
   {\tt\small pranavs108@ucla.edu, jacklishufan@cs.ucla.edu} \\
   {\tt\small siyanz@ucla.edu, adityag@cs.ucla.edu} \\[0.05in]
   {\scriptsize $^*$Equal Contribution}
}

\begin{document}

\twocolumn[
\vbox{
\maketitle
\centering
\includegraphics[width= 0.85\textwidth]{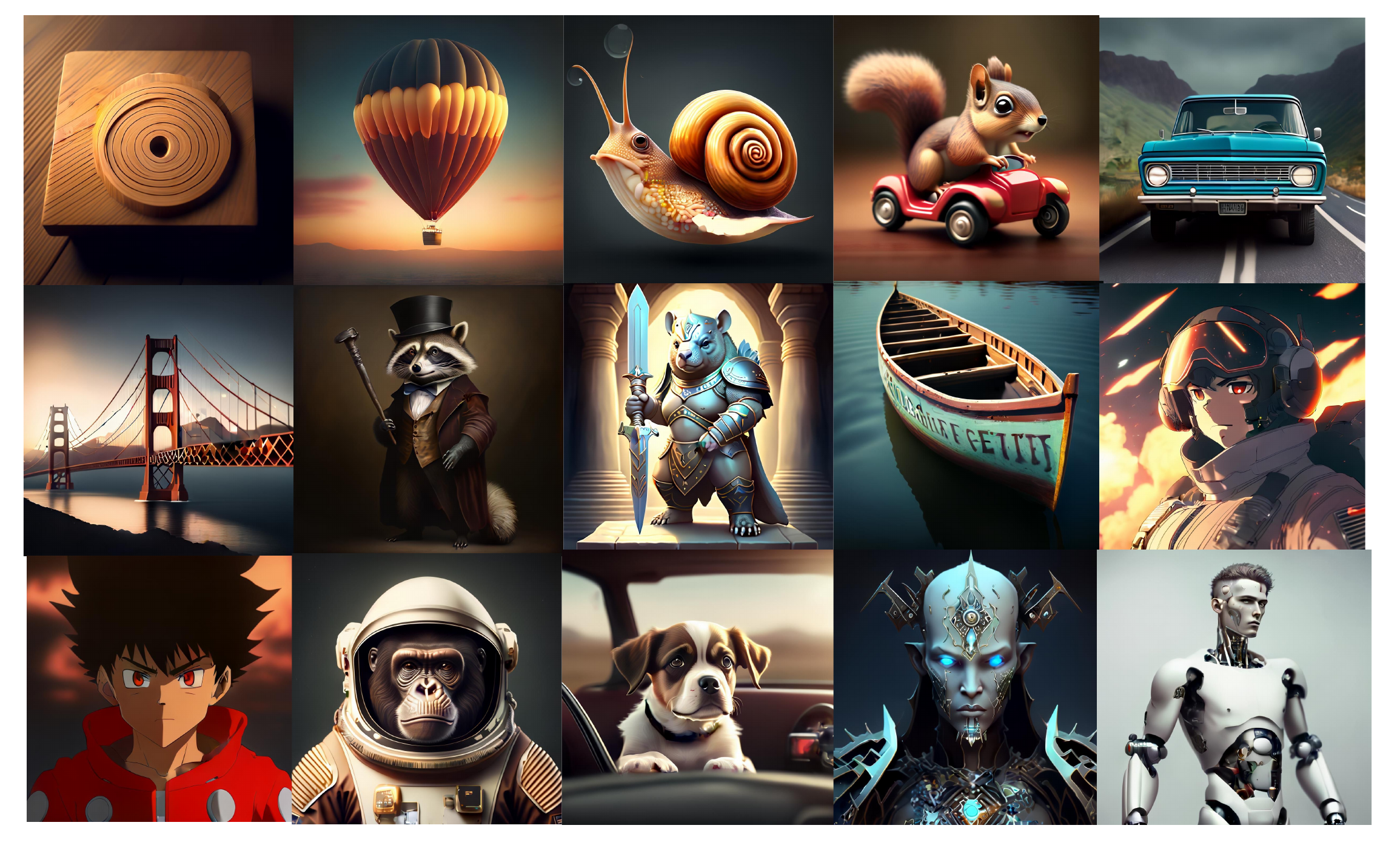}
\captionof{figure}{Speed-RL is a framework for masked generative models to generate high quality images while using significantly fewer steps. Specifically, Speed-RL uses reinforcement learning to optimize a speed and quality reward.}
\label{fig:teaser}
\vspace{1em}}
]

\begin{abstract}
Masked Generative Models (MGM)s demonstrate strong capabilities in generating high-fidelity images. However, they need many sampling steps to create high-quality generations, resulting in slow inference speed. In this work, we propose Speed-RL, a novel paradigm for accelerating a pretrained MGMs to generate high-quality images in fewer steps. Unlike conventional distillation methods which formulate the acceleration problem as a distribution matching problem, where a few-step student model is trained to match the distribution generated by a many-step teacher model, we consider this problem as a reinforcement learning problem. Since the goal of acceleration is to generate high quality images in fewer steps, we can combine a quality reward with a speed reward and finetune the base model using reinforcement learning with the combined reward as the optimization target. Through extensive experiments, we show that the proposed method was able to accelerate the base model by a factor of 3x while maintaining comparable image quality. 
\end{abstract} 
\section{Introduction}
Diffusion models have been shown to generate high-quality images from text prompts \cite{podell2023sdxl, rombach2022high, xie2025sana, pmlr-v139-ramesh21a}. A new framework, Masked generative models (MGMs), has recently emerged as a powerful class of models for a wide range of generative tasks \cite{chang2022maskgit, austin2021structured, shi2024simplified, hu2024mask}. MGMs have been widely applied in producing high-quality image generation \cite{bai2024meissonic, chang2023muse, patil2024amused, hu2024mask, chang2022maskgit}. They have also been successfully applied to video generation \cite{susladkar2024motionaura, fuest2025maskflow} and text generation \cite{nie2025large, lou2023discrete, gong2024scaling, deschenaux2024beyond}. Despite these strengths, MGMs typically require many sampling steps to produce high quality images, which makes real-time or low-latency applications impractical. Therefore, reducing the number of sampling steps presents an impactful opportunity. If we can preserve generation quality while cutting inference cost, MGMs become far more useful in production settings ranging from interactive content creation to on-device synthesis.

Most prior work on accelerating generative diffusion models frames the problem as a distillation or distribution-matching problem. These methods train a compact, few-step student to emulate a many-step teacher While effective in some settings, these approaches are limited by the need to explicitly model the teacher's distribution and by instability when the student's sampling trajectories diverge from those seen during teacher training. For discrete, high entropy token distributions typical of masked image generators, sample-based estimators for divergence (e.g. naive KL estimators) suffer from high variance, which can destabilize training and limit the potential benefits of direct distillation.

In parallel, reinforcement learning (RL) has recently been used to fine-tune generative models by directly optimizing human preference or quality-based rewards \cite{zhang2024large, kordzanganeh2024pixel}. These works suggest that the direct optimization quality and inference cost via RL is a promising direction for high performance generation.

In this work, we take a different view. That is, accelerating a pretrained MGM by directly optimizing for the downstream objective of high image quality and low sampling cost using reinforcement learning. We introduce Speed-RL, a framework that combines a task-level quality reward (e.g. ImageReward \cite{xu2023imagereward}, HPSv2 \cite{wu2023human}, CLIP \cite{radford2021learning}) with an explicit speed reward that favors shorter sampling trajectories. To make policy optimization stable in the discrete, high entropy setting of MGMs, we adapt group-ratio policy optimization (GRPO) \cite{shao2024deepseekmath} with a low-variance KL regularizer and a cross-entropy based estimator that more faithfully captures the full token distribution rather than relying only on sampled tokens. This combination produces stable, directed fine-tuning of the base model toward concise, high quality generation trajectories.

We conduct extensive experiments to test the effectiveness of Speed-RL. We show that Speed-RL can accelerate the base model by a factor of 3x while maintaining perceptual and reward-driven quality metrics.

Our main contributions include:
\begin{itemize}
    \item We cast the problem of accelerating discrete masked diffusion models as a reinforcement learning problem and propose a combined quality + speed reward that directly optimizes the practical objective of interest.
    \item We adapt GRPO for high entropy discrete generators by introducing a low-variance KL estimator based on cross-entropy terms, improving training stability for MGMs.
    \item We empirically validate Speed-RL on a 1B-parameter Meissonic base model \cite{bai2024meissonic} and show that it achieves up to 3x inference speedups with comparable ImageReward, HPSv2.1, and CLIP scores. We include extensive ablation studies that isolate the effects of different hyperparameters.
\end{itemize}

\section{Related Works}

\subsection{Diffusion Model Acceleration}
Diffusion models have demonstrated remarkable generative performance across modalities, but their iterative sampling process leads to slow inference. Prior works have sought to accelerate diffusion-based generation through distillation and model compression \cite{salimans2022progressive, song2023consistency}. Other works reformulate diffusion as flow matching to reduce the number of integration steps \cite{liu2022flow, dao2025self}. However, these approaches depend on accurate distribution matching between teacher and student models, which can be unstable. This is particularly true for discrete masked diffusion models where high entropy token distributions make divergence estimation noisy.

\subsection{Masked Generative Models}
MGMs such as MaskGIT \cite{chang2022maskgit} and MUSE \cite{chang2023muse} replace the continuous denoising process of diffusion with discrete token prediction. This leads to parallel decoding and improved sample quality. Subsequent works like Meissonic \cite{bai2024meissonic} and MaskDiT \cite{zheng2023fast} extended these models with diffusion-style iterative unmasking, achieving strong text-to-image fidelity. Despite their efficiency relative to standard diffusion, MGMs still require many sampling steps to reach optimal quality. This motivates our work on direct optimization of speed-quality tradeoffs.

\subsection{Reinforcement Learning for Generative Models}
Recent advances in RL fine-tuning of generative models have shown that non-differentiable quality metrics, such as human preference, can guide large model behavior beyond likelihood training. In text generation, PPO \cite{schulman2017proximalpolicyoptimizationalgorithms} and GRPO \cite{shao2024deepseekmath} have been used to align model outputs with human evaluations. In image generation, ImageReward \cite{xu2023imagereward} and HPSv2.1 \cite{wu2023human} introduced learned aesthetic and preference rewards for text-to-image models. These methods suggest that RL can be used not only for preference alignment but also for task-specific objectives (e.g. inference efficiency).

\subsection{Reinforcement Learning for Diffusion Model Optimization}
Recent works on using RL to fine-tune diffusion models can be grouped into three broad themes: alignment and preference optimization \cite{black2024trainingdiffusionmodelsreinforcement, hu2025dfusiondirectpreferenceoptimization, lamba2025alignmentsafetydiffusionmodels}, efficiency and inference-speed optimization \cite{zhao2025addingconditionalcontroldiffusion, ma2025efficientonlinereinforcementlearning}, and structural/control applications where diffusion models serve in RL or planning roles \cite{dong2025maximumentropyreinforcementlearning, zhu2024diffusionmodelsreinforcementlearning}. Unlike prior RL works that focus on alignment or adapting the generative process for control, we make inference cost (step count and latency) an explicit optimization objective and directly reward shorter, high quality generation trajectories. Furthermore, we jointly optimize architecture-level decisions (step schedules, masking policy) together with reward design and introduce stable RL adaptations for high entropy, discrete token generators. This allows for significant reduction in sampling steps that prior methods do not target.

\section{Method}

We formulate the acceleration problem of discrete diffusion model as a reinforcement learning problem. Given a prompt $c$, a sampled diffusion trajectories $\{x_t\}=\{x_T,x_{T-1}..x_0\}$ from a masked diffusion model(MDM) $p_\theta$, with $x_t$ being a fully masked image and $x_0$  a clean image, our goal is to solve the following optimization objective

\begin{equation}
    \max \mathbb{E}_{x_0\sim p_\theta(x_0|c),c\sim \mathcal{D}}[R(x_0,c)+\frac{\tau}{|\{x_t\}|}]
\end{equation}

where $R(x_0,c)$ is a real-valued reward function computed over pairs of prompts and generative images,$\{x_t\}$ is the number of sampling steps, and  $\tau$ is a scaling factor. The term $\frac{\tau}{|\{x_t\}|}$ serves as a ``speed reward" that increases as the number of sampling steps $\{x_t\}$ decreases. The hyperparameter $\tau$ controls the relative contribution of the speed reward to the whole learning process.

\subsection{Improving GRPO's Stability}.
\label{sec:stability}
We employ GRPO as our reinforcement learning framework. However, there are some unique challenges in applying it to masked image generators. Recall that the standard GRPO contains a KL penalty term computed using the following estimator:  

\begin{equation}
\mathbb{D}_{\mathrm{KL}}\!\left[p_\theta \,\|\, p_{\mathrm{ref}}\right]
=  
\frac{p_{\mathrm{ref}}(x_i \mid c)}{p_{\theta}(x_i\mid c)}
- 
\log \frac{p_{\mathrm{ref}}(x_i \mid c)}{p_{\theta}(x_i  \mid c)}
- 1,
\end{equation}

over a group of generated samples $x_1,...x_G$. However, this estimator only uses actual sampled tokens instead of the whole token distribution produced by $p_{\text{ref}}$. In discrete image generators, $p_{\text{ref}}$ typically has higher entropy than language tasks, with more high likelihood candidates. Using only sampled token in this case leads to high variance and potentially instability in the tail region of the distribution. To address this, we can replace the estimator based on the relation

\begin{equation}
\mathbb{D}_{\mathrm{KL}}\!\left[p_\theta \,\|\, p_{\mathrm{ref}}\right]
=  
\text{H}(p_{\mathrm{ref}}|p_\theta)-H(p_{\mathrm{ref}})
\end{equation}
where $\text{H}(p_\theta|p_{\mathrm{ref}})$ is the cross entropy and $H(p_{\mathrm{ref}})$ is the entropy.

With this adaptation, we can use the following final learning objective in place of GRPO loss


\begin{align}
\mathcal{L}_{\text{Speed-RL}}= A_i \, L_{\mathrm{ce}}\!\left(\mathbf{x_i},\, p_\theta(X_i \mid c_i)\right)
& \nonumber \\
 \quad +\;
\beta L_{\mathrm{ce}}( p_{\mathrm{ref}}(X_i&  \mid c_i),\, p_\theta(X_i \mid c_i))
\label{eq:loss-speed}
\end{align}

where $A_i$ is the group advantage term in GRPO, $L_{\mathrm{ce}}$ is the cross entropy loss, $p(X_i \mid c_i)$ are per-token probability vector and  $\mathbf{x_i}$ is a one-hot vector. We will show that this approach result in the same gradient as the vanilla GRPO loss with our proposed KL divergence estimator. We will provide a full proof in the appendix.



\subsection{Low Quality Sample Filtering}
We employ a technique to filter out prompts where the reward model shows high variance and uncertainty in its scoring. The overall method is as follows. First, we generate multiple samples for each prompt. Then, we score each sample with a reward model. Next, we calculate the standard deviation of the reward scores for that prompt, and then filter out prompts where the standard deviation is greater than a threshold. 

More formally, 
The filtering mechanism detects and rejects "low-diversity" batches where generated samples have similar quality (low variance in rewards). Such batches provide weak learning signals because:
\begin{itemize}
    \item Weak advantage signals: GRPO relies on reward differences to compute advantages. If all samples have similar rewards, advantages become noisy.
    \item Poor exploration: Low variance indicates the model is producing repetitive outputs, limiting exploration of the reward landscape.
    \item Inefficient learning: Training on uniform-quality batches wastes compute without teaching the model meaningful distinctions.
\end{itemize}

\subsubsection*{Step 1: Batch Generation and Reward Computation}
For a given prompt, the system generates a batch of images with varying inference steps:
\begin{itemize}
    \item Images: $\{x_1, x_2, \dots, x_n\}$ where $n = \text{batch\_size}$
    \item Rewards: $R = \{r_1, r_2, \dots, r_n\}$ where $r_i = f_{\text{reward}}(x_i, \text{prompt})$
\end{itemize}
The final score includes speed rewards:
$$r_i = r_{\text{base}}(x_i) + \alpha \cdot \left(\frac{8}{\text{nfe}_i}\right)$$
where $\alpha$ is $\text{speed\_reward\_factor}$ and $\text{nfe}_i$ is the number of function evaluations for sample $i$.

\subsubsection*{Step 2: Standard Deviation Calculation}
For multi-GPU training, rewards are gathered across all GPUs:
$$R_{\text{all}} = [R^{(0)}, R^{(1)}, \dots, R^{(k-1)}] \quad \% k \text{ GPUs}$$
The reward standard deviation is computed:
$$\sigma_{\text{batch}} = \text{mean}(\{\text{std}(R^{(0)}), \text{std}(R^{(1)}), \dots, \text{std}(R^{(k-1)})\})$$

\subsubsection*{Step 3: Historical Tracking and Percentile Threshold}
The system maintains a sliding window of historical standard deviations:
$$H = \{\sigma_1, \sigma_2, \dots, \sigma_m\} \text{ where } m \le \text{history\_size}$$
Default: $\text{history\_size} = 100$

The acceptance threshold is the $p$-th percentile of historical values:
$$\tau = \text{percentile}(H, p)$$
Default: $p = 10$ (10th percentile - only batches with std dev in the bottom 10\% are rejected)


\subsubsection*{Step 4: Accept/Reject Decision}
The batch is accepted if:
$$\sigma_{\text{batch}} \ge \tau \quad \text{OR} \quad \text{resample\_attempts} > \text{max\_attempts}$$
Otherwise, regenerate the entire batch and recompute.

\subsubsection*{Step 5: History Update}
After acceptance, update the sliding window:
$$H \leftarrow H \cup \{\sigma_{\text{batch}}\}$$
if $|H| > \text{history\_size}$:
$$H \leftarrow H[1:] \quad \% \text{ Remove oldest}$$

In this way, we keep only low variance prompts for training. Overall, this can help identify problematic prompts since high standard deviation indicates the reward model is uncertain about how to score that prompt. We can improve the training data quality by removing these prompts. We see more consistent and reliable reward signals when evaluating models after using this technique. This design is inspired by VCRL which uses this for language \cite{jiang2025vcrl}.

\section{Experiment Results}

\subsection{Setup}
\subsubsection{Dataset}
We sample 100k prompts from LAION-Aesthetics \cite{schuhmann2022laion5bopenlargescaledataset}, BLIP3O-60k \cite{chen2025blip3}

\subsubsection{Reward Models}
We set up a pipeline to use Meissonic \cite{bai2024meissonic}, a foundational text-to-image MGM with 1B parameters, to generate images for evaluating different reward model options. We sample 20 prompts from BLIP3o-60k and generate 100 images each using 16, 32, and 48 sampling steps. We then run the ImageReward, PickScore, HPSv2.1, and CLIP on each of these generated images. We visually inspect the top-k scored images for different prompts for each reward model and calculate average scores and standard deviations for each reward model across steps. Based on this analysis, we decide that for a given reward model, an image generated by our framework is high quality if it gets a score greater than or equal to one standard deviation above the average score for the baseline model's generations. We establish that an image generated by our framework is low quality if it gets a score below the average score for the baseline model's generations. Overall, this helps us evaluate the quality of different reward models and determine thresholds for high quality and low quality images generated by our framework.

\subsubsection{Training}
We train the base model using our modified GRPO objective. We use a global batch size of 96 and across 8 GPUs and a learning rate of $5\times 10^-7$ for 1500 steps.

\subsection{Qualitative Results} 
\begin{figure*}
    \centering
    \includegraphics[width=1\linewidth]{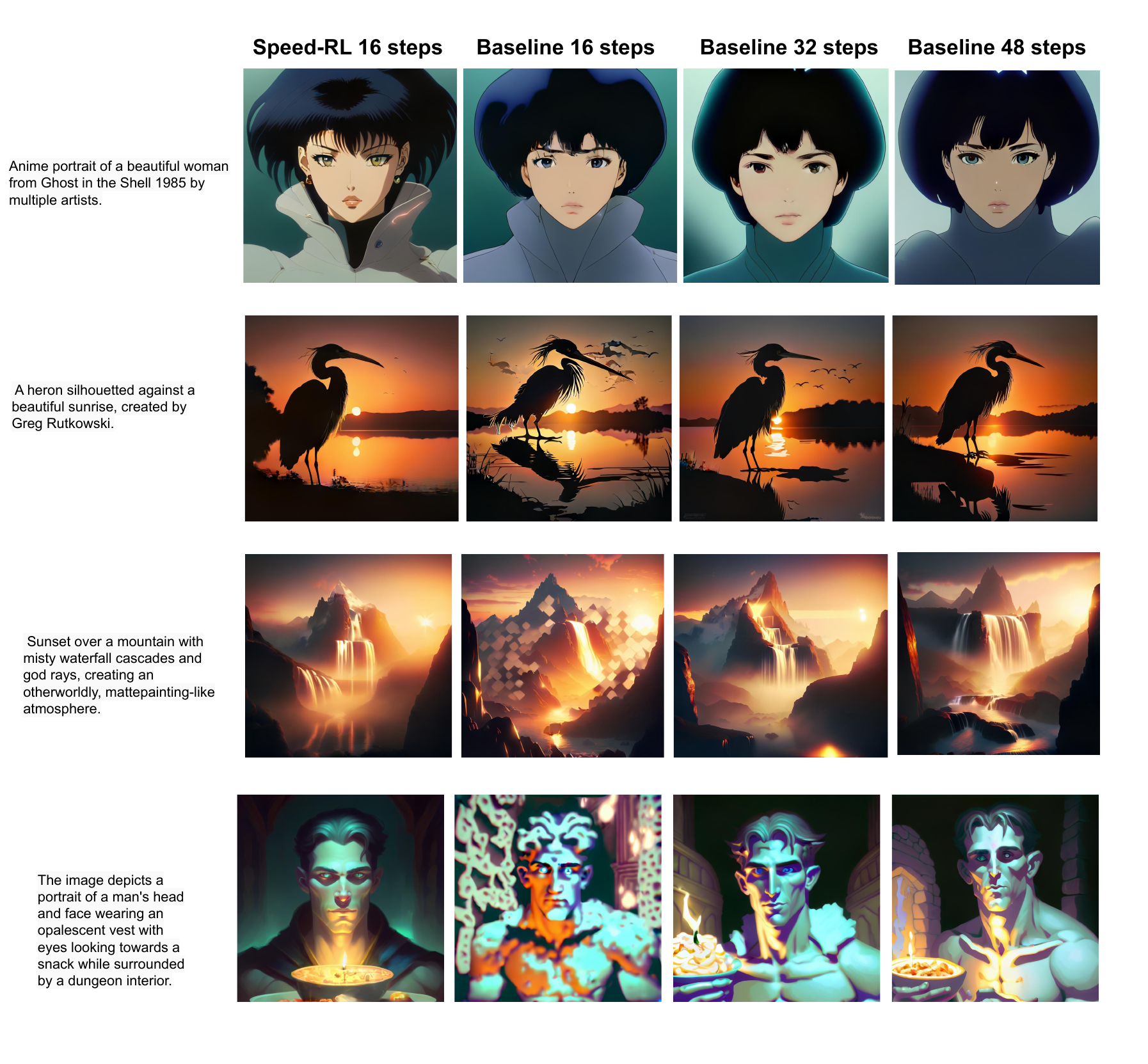}
    \caption{\textbf{Side-by-side qualitative comparison of Speed-RL and Meissonic (baseline)}. Speed-RL produces higher quality images while using significantly fewer steps.}
    \label{fig:side}
\end{figure*}

We report qualitative comparisons in \cref{fig:side}. We compare text-to-image generations of Speed-RL with 16 sampling steps  and compare it with generations from baseline model Meissonic at 16, 32, and 48 steps on HPSv2.1 dataset. These visual results demonstrate that Speed-RL was able to achieve comparable generation quality with a $4\times$ speedup. 

\subsection{Main Results}
We report results on the HPSv2.1 benchmark and the PartiPrompts benchmark in Table 1. We employ ImageReward, HPSv2.1, and CLIP PickScore score as our quality reward models. We compare these metrics of the base model and finetuned model under different sampling steps. Speed-RL delivers quality on par with the baseline across both benchmarks and all three reward models, while achieving a 3x speedup, requiring only one-third of the sampling steps used by the baseline.

\begin{table*}[t]
\centering
\renewcommand{\arraystretch}{1.5}
\label{tab:merged-benchmarks}

\begin{tabular}{l c c c  |  l c c c}
\hline
\multicolumn{4}{c|}{\textbf{HPS Benchmark}} &
\multicolumn{4}{c}{\textbf{PartiPrompts Benchmark}} \\
\hline
\textbf{Model} & \textbf{Steps} & \textbf{ImageReward} & \textbf{PickScore} &
\textbf{Model} & \textbf{Steps} & \textbf{ImageReward} & \textbf{PickScore} \\
\hline

Meissonic & 48 & 41.901 & 21.416  &
Meissonic & 48 & 28.903 & 21.498 \\

\hline
Speed-RL & \textbf{16} & \textbf{63.346} & \textbf{21.607} &
Speed-RL & \textbf{16} & \textbf{45.947} & \textbf{21.394} \\

Speed-RL & 32 & 69.019 & 21.682 &
Speed-RL & 32 & 43.349 & 21.414 \\

Speed-RL & 48 & 64.485 & 21.611 &
Speed-RL & 48 & 34.522 & 21.335 \\
\hline
\end{tabular}
\caption{Comparison of Meissonic and Speed-RL across HPSv2.1 and PartiPrompts benchmarks. Across both benchmarks and all reward models, Speed-RL matches baseline quality and achieves a 3× speedup using only one-third of the sampling steps.}
\end{table*}

\begin{table*}[htbp]
\centering
\renewcommand{\arraystretch}{1.5}
\label{tab:hpsv2-scores}
\begin{tabular}{llccccc}
\hline
\textbf{Model} & \textbf{Step} & \textbf{Anime} & \textbf{Concept Art} & \textbf{Paintings} & \textbf{Photo} & \textbf{Overall} \\
\hline
Baseline & 48 & 27.469 & 25.357 & 25.248 & 22.835 & 25.227 \\
\hline
Speed-RL & \textbf{16} & \textbf{28.457} & \textbf{27.849} & \textbf{26.581} & \textbf{24.491} & \textbf{26.845}\\
Speed-RL & 32 & 29.169 & 27.411 & 26.926 & 24.701 & 27.052\\
Speed-RL & 48 & 28.821 & 27.139 & 26.741 & 24.551 & 26.813\\
\hline
\end{tabular}
\caption{HPSv2.1 Scores Comparison (Speed-RL vs. Baseline) by Category. Speed-RL's overall HPSv2.1 score is +1.618 than the baseline model's score while using only 16 steps compared to 48. The ConceptArt and Photo categories show the greatest improvement.}
\end{table*}

\begin{table*}[htbp]
\centering
\renewcommand{\arraystretch}{1.5}
\label{tab:parti-hpsv2-scores}
\resizebox{1.0\linewidth}{!}{ 
\begin{tabular}{ll|ccccccccc}
\hline
\textbf{Model} & \textbf{Step} & \textbf{Abstract} & \textbf{Animals} & \textbf{Artifacts} & \textbf{Arts} & \textbf{Food \& Bev.} & \textbf{Illustrations} & \textbf{Indoor} & \textbf{Outdoor} & \textbf{Overall} \\
\hline
Baseline & 48 & 18.210 & 28.239 & 24.528 & 25.922 & 22.165 & 24.703 & 23.410 & 23.137 & 23.763 \\
\hline
Speed-RL & \textbf{16} & \textbf{22.329} & \textbf{28.895} & \textbf{24.459} & \textbf{27.887} & \textbf{21.987} & \textbf{24.756} & \textbf{25.294} & \textbf{24.669} & \textbf{22.252} \\
Speed-RL & 32 & 21.551 & 29.448 & 24.349 & 27.901 & 20.977 & 24.548 & 25.793 & 24.937 & 24.888 \\
Speed-RL & 48 & 21.017 & 29.083 & 23.949 & 28.013 & 20.389 & 24.224 & 25.322 & 24.258 & 24.488 \\
\hline
\end{tabular}}
\caption{PartiPrompts Benchmark HPSv2.1 Scores (Speed-RL vs. Baseline) by Category. Speed-RL achieves a very comparable HPSv2.1 reward score with \textbf{3x} fewer steps than the baseline.}
\end{table*}

These results show that with RL finetuning, we can obtain higher image quality than the base model while also achieving a speedup of 3x.  We observe that the improvements in low-sampling steps is more significant, which may be an effect of our speed reward that focus the positive learning signal on generations with fewer steps.

\subsection{Ablation Studies}

We measure the effect of various design choices by conducting thorough ablation studies on Speed-RL. 

\subsubsection{Speed Reward and KL Regularizer}

\begin{figure}[h]
    \centering
    \includegraphics[width=0.99\linewidth]{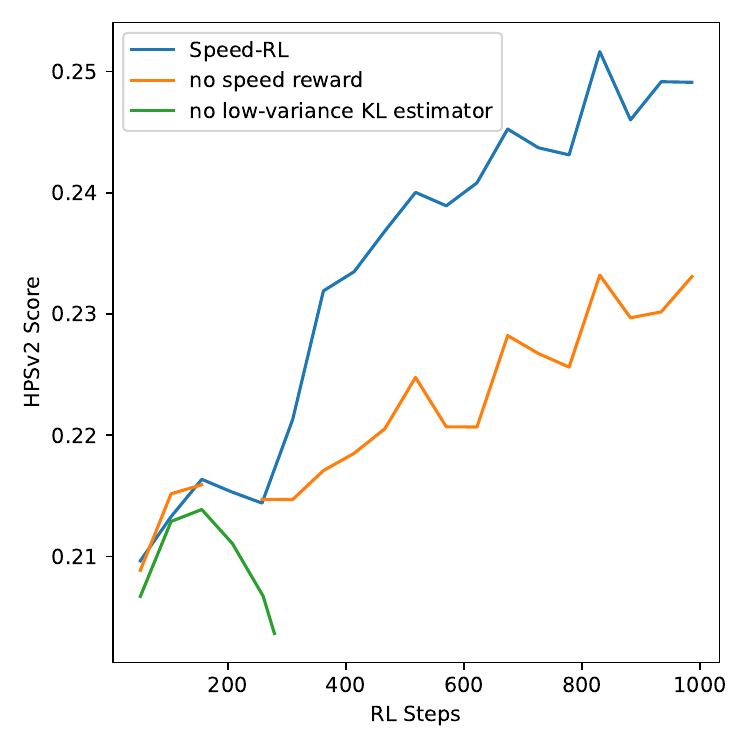}
    \caption{Ablation study on our speed reward and KL regularizer. We see a divergence in HPSv2.1 scores after 300 RL steps, where Speed-RL increases faster compared to having no speed reward and no low-variance KL estimator.}
    \label{fig:ablation}
\end{figure}

In Figure \ref{fig:ablation}, we visualize the average reward of generated samples throughout the training process on 16-step generations. We find that without speed reward, the training converges slower as less positive advantages are assigned to samples with fewer steps. Without our proposed change to the KL regularizer, the training diverges, highlighting the instability problems with vanilla sample-based KL estimator.

\subsubsection{Enabling Per Token Likelihood}

\begin{figure}[h]
    \centering
    \includegraphics[width=0.99\linewidth]{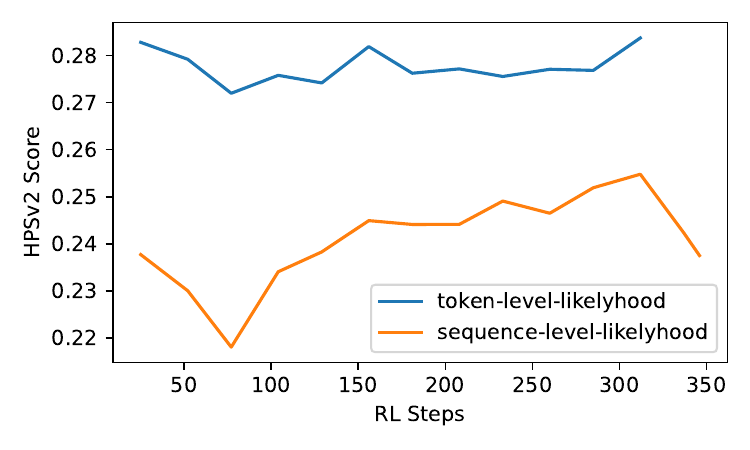}
    \caption{Ablation study on the effect of enabling per-token-likelihood. There is a gap between using token-level-likelihood and sequence-level-likelihood where using token-level-likelihood consistently has a HPSv2.1 score higher than using sequence-level-likelihood.}
    \label{fig:ablation1}
\end{figure}

We experiment with both including and not including per-token-likelihood where all other hyperparameters are held constant. We observe that not including per-token-likelihood results in a higher average HPSv2.1 reward score across over 300 steps and thus we do not include per-token-likelihood in our optimal selection of hyperparameter values (Figure \ref{fig:ablation1}).

\begin{figure}[h]
    \centering
    \includegraphics[width=0.99\linewidth]{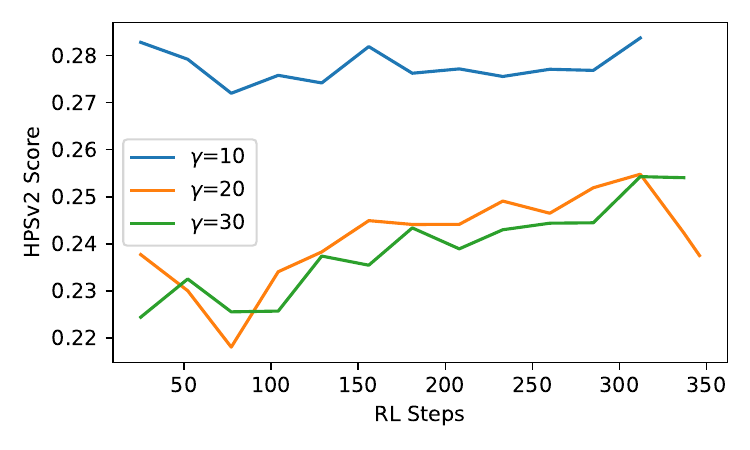}
    \caption{Ablation study on varying the reward standard deviation percentile threshold.}
    \label{fig:ablation_ga,a}
\end{figure}

\subsubsection{Low Quality Sample Filtering}
We try a low-quality sample filtering method. This tracks the historical standard deviation of rewards within each group of generations, and if a newly generated group has a reward standard deviation below a certain percentile threshold, it discards and resamples that entire group. We measure the effect of different percentile threshold values for this method, and observe that as the threshold increases, the average HPSv2.1 reward becomes increasingly noisy. We choose 10.0 percent for our threshold value to strike a balance between very high noise and filtering too few groups.

\begin{figure}[h]
    \centering
    \includegraphics[width=0.99\linewidth]{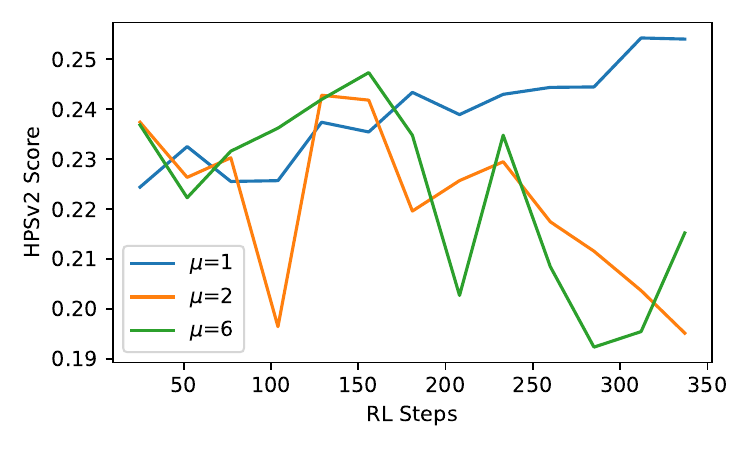}
    \caption{Ablation study on varying the number of gradient update steps per batch.}
    \label{fig:ablation-num-iterations}
\end{figure}

\subsubsection{Number Of Gradient Update Steps Per Batch}
We vary the number of gradient update steps per batch of generated images. A higher value results in better sample efficiency and fewer image generation calls (faster training, less compute). However, this means optimizing on "stale" samples, i.e. the policy updates but keep training on old generations. This also has a risk of overfitting. On the other hand, lower values result in more frequent image generation with updated policy (fresher samples). This means better alignment between current policy and training data at the cost of slower training overall (more compute for generation). We find that setting this value to 1 at the cost of slower training time gets the best performance.

\subsubsection{CFG Value}
We try different Classifier-Free Guidance (CFG) values and analyze the overall variance and average reward. We prefer relatively higher CFG values to get more diversity in generation.

\begin{figure}[h]
    \centering
    \includegraphics[width=0.99\linewidth]{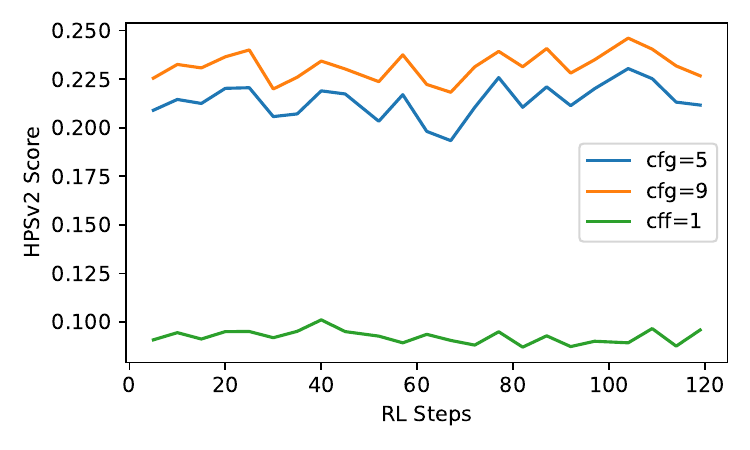}
    \caption{Ablation study on varying the Classifier-Free Guidance (CFG) value.}
    \label{fig:ablation-cfg}
\end{figure}

\section{Conclusion}

In this work, we introduced Speed-RL, a new framework to accelerate masked image generators that moves beyond distillation and directly optimizes for high quality images produces in minimal sampling steps. By pairing a speed reward with a quality reward and introducing a low variance KL regularizer tailored to MGMs, Speed-RL delivers 3x inference acceleration with no degradation in ImageReward, HPSv2.1, or CLIP metrics across two benchmarks. Our work shows that RL is not only feasible but advantageous for controlling the sampling behavior of discrete diffusion-style models. We believe this represents a promising direction for scalable generative modeling with efficient inference.

\clearpage
{
    \small
    \bibliographystyle{ieeenat_fullname}
    \bibliography{main}

@String(AAAI = {AAAI})

@inproceedings{chang2022maskgit,
  title={Maskgit: Masked generative image transformer},
  author={Chang, Huiwen and Zhang, Han and Jiang, Lu and Liu, Ce and Freeman, William T},
  booktitle={Proceedings of the IEEE/CVF conference on computer vision and pattern recognition},
  pages={11315--11325},
  year={2022}
}

@inproceedings{zhang2024large,
  title={Large-scale reinforcement learning for diffusion models},
  author={Zhang, Yinan and Tzeng, Eric and Du, Yilun and Kislyuk, Dmitry},
  booktitle={European Conference on Computer Vision},
  pages={1--17},
  year={2024},
  organization={Springer}
}

@article{kordzanganeh2024pixel,
  title={Pixel-wise RL on Diffusion Models: Reinforcement Learning from Rich Feedback},
  author={Kordzanganeh, Mo and Keshvary, Danial and Arian, Nariman},
  journal={arXiv preprint arXiv:2404.04356},
  year={2024}
}

@inproceedings{bai2024meissonic,
  title={Meissonic: Revitalizing masked generative transformers for efficient high-resolution text-to-image synthesis},
  author={Bai, Jinbin and Ye, Tian and Chow, Wei and Song, Enxin and Chen, Qing-Guo and Li, Xiangtai and Dong, Zhen and Zhu, Lei and Yan, Shuicheng},
  booktitle={The Thirteenth International Conference on Learning Representations},
  year={2024}
}

@article{xu2023imagereward,
  title={Imagereward: Learning and evaluating human preferences for text-to-image generation},
  author={Xu, Jiazheng and Liu, Xiao and Wu, Yuchen and Tong, Yuxuan and Li, Qinkai and Ding, Ming and Tang, Jie and Dong, Yuxiao},
  journal={Advances in Neural Information Processing Systems},
  volume={36},
  pages={15903--15935},
  year={2023}
}

@article{wu2023human,
  title={Human preference score v2: A solid benchmark for evaluating human preferences of text-to-image synthesis},
  author={Wu, Xiaoshi and Hao, Yiming and Sun, Keqiang and Chen, Yixiong and Zhu, Feng and Zhao, Rui and Li, Hongsheng},
  journal={arXiv preprint arXiv:2306.09341},
  year={2023}
}

@inproceedings{radford2021learning,
  title={Learning transferable visual models from natural language supervision},
  author={Radford, Alec and Kim, Jong Wook and Hallacy, Chris and Ramesh, Aditya and Goh, Gabriel and Agarwal, Sandhini and Sastry, Girish and Askell, Amanda and Mishkin, Pamela and Clark, Jack and others},
  booktitle={International conference on machine learning},
  pages={8748--8763},
  year={2021},
  organization={PmLR}
}

@article{shao2024deepseekmath,
  title={Deepseekmath: Pushing the limits of mathematical reasoning in open language models},
  author={Shao, Zhihong and Wang, Peiyi and Zhu, Qihao and Xu, Runxin and Song, Junxiao and Bi, Xiao and Zhang, Haowei and Zhang, Mingchuan and Li, YK and Wu, Yang and others},
  journal={arXiv preprint arXiv:2402.03300},
  year={2024}
}

@article{salimans2022progressive,
  title={Progressive distillation for fast sampling of diffusion models},
  author={Salimans, Tim and Ho, Jonathan},
  journal={arXiv preprint arXiv:2202.00512},
  year={2022}
}

@article{song2023consistency,
  title={Consistency models},
  author={Song, Yang and Dhariwal, Prafulla and Chen, Mark and Sutskever, Ilya},
  year={2023}
}

@article{liu2022flow,
  title={Flow straight and fast: Learning to generate and transfer data with rectified flow},
  author={Liu, Xingchao and Gong, Chengyue and Liu, Qiang},
  journal={arXiv preprint arXiv:2209.03003},
  year={2022}
}

@inproceedings{dao2025self,
  title={Self-Corrected Flow Distillation for Consistent One-Step and Few-Step Image Generation},
  author={Dao, Quan and Phung, Hao and Dao, Trung Tuan and Metaxas, Dimitris N and Tran, Anh},
  booktitle={Proceedings of the AAAI Conference on Artificial Intelligence},
  volume={39},
  number={3},
  pages={2654--2662},
  year={2025}
}

@article{chang2023muse,
  title={Muse: Text-to-image generation via masked generative transformers},
  author={Chang, Huiwen and Zhang, Han and Barber, Jarred and Maschinot, AJ and Lezama, Jose and Jiang, Lu and Yang, Ming-Hsuan and Murphy, Kevin and Freeman, William T and Rubinstein, Michael and others},
  journal={arXiv preprint arXiv:2301.00704},
  year={2023}
}

@article{zheng2023fast,
  title={Fast training of diffusion models with masked transformers},
  author={Zheng, Hongkai and Nie, Weili and Vahdat, Arash and Anandkumar, Anima},
  journal={arXiv preprint arXiv:2306.09305},
  year={2023}
}

@misc{schulman2017proximalpolicyoptimizationalgorithms,
      title={Proximal Policy Optimization Algorithms}, 
      author={John Schulman and Filip Wolski and Prafulla Dhariwal and Alec Radford and Oleg Klimov},
      year={2017},
      eprint={1707.06347},
      archivePrefix={arXiv},
      primaryClass={cs.LG},
      url={https://arxiv.org/abs/1707.06347}, 
}

@misc{black2024trainingdiffusionmodelsreinforcement,
      title={Training Diffusion Models with Reinforcement Learning}, 
      author={Kevin Black and Michael Janner and Yilun Du and Ilya Kostrikov and Sergey Levine},
      year={2024},
      eprint={2305.13301},
      archivePrefix={arXiv},
      primaryClass={cs.LG},
      url={https://arxiv.org/abs/2305.13301}, 
}

@misc{zhao2025addingconditionalcontroldiffusion,
      title={Adding Conditional Control to Diffusion Models with Reinforcement Learning}, 
      author={Yulai Zhao and Masatoshi Uehara and Gabriele Scalia and Sunyuan Kung and Tommaso Biancalani and Sergey Levine and Ehsan Hajiramezanali},
      year={2025},
      eprint={2406.12120},
      archivePrefix={arXiv},
      primaryClass={cs.LG},
      url={https://arxiv.org/abs/2406.12120}, 
}

@misc{hu2025dfusiondirectpreferenceoptimization,
      title={D-Fusion: Direct Preference Optimization for Aligning Diffusion Models with Visually Consistent Samples}, 
      author={Zijing Hu and Fengda Zhang and Kun Kuang},
      year={2025},
      eprint={2505.22002},
      archivePrefix={arXiv},
      primaryClass={cs.CV},
      url={https://arxiv.org/abs/2505.22002}, 
}

@misc{lamba2025alignmentsafetydiffusionmodels,
      title={Alignment and Safety of Diffusion Models via Reinforcement Learning and Reward Modeling: A Survey}, 
      author={Preeti Lamba and Kiran Ravish and Ankita Kushwaha and Pawan Kumar},
      year={2025},
      eprint={2505.17352},
      archivePrefix={arXiv},
      primaryClass={cs.CV},
      url={https://arxiv.org/abs/2505.17352}, 
}

@misc{ma2025efficientonlinereinforcementlearning,
      title={Efficient Online Reinforcement Learning for Diffusion Policy}, 
      author={Haitong Ma and Tianyi Chen and Kai Wang and Na Li and Bo Dai},
      year={2025},
      eprint={2502.00361},
      archivePrefix={arXiv},
      primaryClass={cs.LG},
      url={https://arxiv.org/abs/2502.00361}, 
}

@misc{dong2025maximumentropyreinforcementlearning,
      title={Maximum Entropy Reinforcement Learning with Diffusion Policy}, 
      author={Xiaoyi Dong and Jian Cheng and Xi Sheryl Zhang},
      year={2025},
      eprint={2502.11612},
      archivePrefix={arXiv},
      primaryClass={cs.LG},
      url={https://arxiv.org/abs/2502.11612}, 
}

@misc{zhu2024diffusionmodelsreinforcementlearning,
      title={Diffusion Models for Reinforcement Learning: A Survey}, 
      author={Zhengbang Zhu and Hanye Zhao and Haoran He and Yichao Zhong and Shenyu Zhang and Haoquan Guo and Tingting Chen and Weinan Zhang},
      year={2024},
      eprint={2311.01223},
      archivePrefix={arXiv},
      primaryClass={cs.LG},
      url={https://arxiv.org/abs/2311.01223}, 
}

@article{jiang2025vcrl,
  title={VCRL: Variance-based Curriculum Reinforcement Learning for Large Language Models},
  author={Jiang, Guochao and Feng, Wenfeng and Quan, Guofeng and Hao, Chuzhan and Zhang, Yuewei and Liu, Guohua and Wang, Hao},
  journal={arXiv preprint arXiv:2509.19803},
  year={2025}
}

@inproceedings{rombach2022high,
  title={High-resolution image synthesis with latent diffusion models},
  author={Rombach, Robin and Blattmann, Andreas and Lorenz, Dominik and Esser, Patrick and Ommer, Bj{\"o}rn},
  booktitle={Proceedings of the IEEE/CVF conference on computer vision and pattern recognition},
  pages={10684--10695},
  year={2022}
}

@article{podell2023sdxl,
  title={Sdxl: Improving latent diffusion models for high-resolution image synthesis},
  author={Podell, Dustin and English, Zion and Lacey, Kyle and Blattmann, Andreas and Dockhorn, Tim and M{\"u}ller, Jonas and Penna, Joe and Rombach, Robin},
  journal={arXiv preprint arXiv:2307.01952},
  year={2023}
}

@article{chen2025blip3,
  title={Blip3-o: A family of fully open unified multimodal models-architecture, training and dataset},
  author={Chen, Jiuhai and Xu, Zhiyang and Pan, Xichen and Hu, Yushi and Qin, Can and Goldstein, Tom and Huang, Lifu and Zhou, Tianyi and Xie, Saining and Savarese, Silvio and others},
  journal={arXiv preprint arXiv:2505.09568},
  year={2025}
}

@misc{schuhmann2022laion5bopenlargescaledataset,
      title={LAION-5B: An open large-scale dataset for training next generation image-text models}, 
      author={Christoph Schuhmann and Romain Beaumont and Richard Vencu and Cade Gordon and Ross Wightman and Mehdi Cherti and Theo Coombes and Aarush Katta and Clayton Mullis and Mitchell Wortsman and Patrick Schramowski and Srivatsa Kundurthy and Katherine Crowson and Ludwig Schmidt and Robert Kaczmarczyk and Jenia Jitsev},
      year={2022},
      eprint={2210.08402},
      archivePrefix={arXiv},
      primaryClass={cs.CV},
      url={https://arxiv.org/abs/2210.08402}, 
}

@misc{sun2023journeydbbenchmarkgenerativeimage,
      title={JourneyDB: A Benchmark for Generative Image Understanding}, 
      author={Keqiang Sun and Junting Pan and Yuying Ge and Hao Li and Haodong Duan and Xiaoshi Wu and Renrui Zhang and Aojun Zhou and Zipeng Qin and Yi Wang and Jifeng Dai and Yu Qiao and Limin Wang and Hongsheng Li},
      year={2023},
      eprint={2307.00716},
      archivePrefix={arXiv},
      primaryClass={cs.CV},
      url={https://arxiv.org/abs/2307.00716}, 
}

@misc{lin2015microsoftcococommonobjects,
      title={Microsoft COCO: Common Objects in Context}, 
      author={Tsung-Yi Lin and Michael Maire and Serge Belongie and Lubomir Bourdev and Ross Girshick and James Hays and Pietro Perona and Deva Ramanan and C. Lawrence Zitnick and Piotr Dollár},
      year={2015},
      eprint={1405.0312},
      archivePrefix={arXiv},
      primaryClass={cs.CV},
      url={https://arxiv.org/abs/1405.0312}, 
}

@article{austin2021structured,
  title={Structured denoising diffusion models in discrete state-spaces},
  author={Austin, Jacob and Johnson, Daniel D and Ho, Jonathan and Tarlow, Daniel and Van Den Berg, Rianne},
  journal={Advances in neural information processing systems},
  volume={34},
  pages={17981--17993},
  year={2021}
}

@article{hu2024mask,
  title={[MASK] is All You Need},
  author={Hu, Vincent Tao and Ommer, Bj{\"o}rn},
  journal={arXiv preprint arXiv:2412.06787},
  year={2024}
}

@article{shi2024simplified,
  title={Simplified and generalized masked diffusion for discrete data},
  author={Shi, Jiaxin and Han, Kehang and Wang, Zhe and Doucet, Arnaud and Titsias, Michalis},
  journal={Advances in neural information processing systems},
  volume={37},
  pages={103131--103167},
  year={2024}
}

@article{patil2024amused,
  title={amused: An open muse reproduction},
  author={Patil, Suraj and Berman, William and Rombach, Robin and von Platen, Patrick},
  journal={arXiv preprint arXiv:2401.01808},
  year={2024}
}

@article{nie2025large,
  title={Large language diffusion models},
  author={Nie, Shen and Zhu, Fengqi and You, Zebin and Zhang, Xiaolu and Ou, Jingyang and Hu, Jun and Zhou, Jun and Lin, Yankai and Wen, Ji-Rong and Li, Chongxuan},
  journal={arXiv preprint arXiv:2502.09992},
  year={2025}
}

@article{lou2023discrete,
  title={Discrete diffusion modeling by estimating the ratios of the data distribution},
  author={Lou, Aaron and Meng, Chenlin and Ermon, Stefano},
  journal={arXiv preprint arXiv:2310.16834},
  year={2023}
}

@article{gong2024scaling,
  title={Scaling diffusion language models via adaptation from autoregressive models},
  author={Gong, Shansan and Agarwal, Shivam and Zhang, Yizhe and Ye, Jiacheng and Zheng, Lin and Li, Mukai and An, Chenxin and Zhao, Peilin and Bi, Wei and Han, Jiawei and others},
  journal={arXiv preprint arXiv:2410.17891},
  year={2024}
}

@article{deschenaux2024beyond,
  title={Beyond autoregression: Fast llms via self-distillation through time},
  author={Deschenaux, Justin and Gulcehre, Caglar},
  journal={arXiv preprint arXiv:2410.21035},
  year={2024}
}

@article{fuest2025maskflow,
  title={Maskflow: Discrete flows for flexible and efficient long video generation},
  author={Fuest, Michael and Hu, Vincent Tao and Ommer, Bj{\"o}rn},
  journal={arXiv preprint arXiv:2502.11234},
  year={2025}
}

@article{susladkar2024motionaura,
  title={MotionAura: Generating High-Quality and Motion Consistent Videos using Discrete Diffusion},
  author={Susladkar, Onkar and Gupta, Jishu Sen and Sehgal, Chirag and Mittal, Sparsh and Singhal, Rekha},
  journal={arXiv preprint arXiv:2410.07659},
  year={2024}
}

@inproceedings{xie2025sana,
  title={SANA: Efficient high-resolution text-to-image synthesis with linear diffusion transformers},
  author={Xie, Enze and Chen, Junsong and Chen, Junyu and Cai, Han and Tang, Haotian and Lin, Yujun and Zhang, Zhekai and Li, Muyang and Zhu, Ligeng and Lu, Yao and others},
  booktitle={The Thirteenth International Conference on Learning Representations},
  year={2025}
}

@InProceedings{pmlr-v139-ramesh21a,
  title = 	 {Zero-Shot Text-to-Image Generation},
  author =       {Ramesh, Aditya and Pavlov, Mikhail and Goh, Gabriel and Gray, Scott and Voss, Chelsea and Radford, Alec and Chen, Mark and Sutskever, Ilya},
  booktitle = 	 {Proceedings of the 38th International Conference on Machine Learning},
  pages = 	 {8821--8831},
  year = 	 {2021},
  editor = 	 {Meila, Marina and Zhang, Tong},
  volume = 	 {139},
  series = 	 {Proceedings of Machine Learning Research},
  month = 	 {18--24 Jul},
  publisher =    {PMLR},
  url = 	 {https://proceedings.mlr.press/v139/ramesh21a.html}
}
}

\clearpage
\appendix
\setcounter{page}{1}
\renewcommand{\thepage}{S\arabic{page}}
\onecolumn

\begin{center}
   \Large \textbf{Supplementary Material}
\end{center}
\vspace{1em}

\section{Additional Experiment Results}

\begin{table*}[htbp]
\centering
\renewcommand{\arraystretch}{1.5}
\label{tab:hpsv2-scores}
\begin{tabular}{llccccc}
\hline
\textbf{Model} & \textbf{Step} & \textbf{Anime} & \textbf{Concept Art} & \textbf{Paintings} & \textbf{Photo} & \textbf{Overall} \\
\hline
Baseline & 16 & 28.401 & 27.508 & 27.539 & 27.149 & 27.649 \\
Baseline & 32 & 28.366 & 27.290 & 27.303 & 26.924 & 27.471 \\
Baseline & 48 & 28.179 & 27.122 & 27.094 & 26.771 & 27.292 \\
\hline
Speed-RL & \textbf{16} & \textbf{28.441} & \textbf{27.643} & \textbf{27.443} & \textbf{27.261} & \textbf{27.697}\\
Speed-RL & 32 & 29.653 & 27.730 & 27.613 & 27.384 & 27.845\\
Speed-RL & 48 & 28.564 & 27.687 & 27.540 & 27.352 & 27.786\\
\hline
\end{tabular}
\caption{HPSv2.0 Scores Comparison (Speed-RL vs. Baseline) by category for the HPS dataset.}
\end{table*}

\begin{table*}[htbp]
\centering
\renewcommand{\arraystretch}{1.5}
\label{tab:parti-hpsv2-scores}
\resizebox{1.0\linewidth}{!}{%
\begin{tabular}{ll|ccccccccc}
\hline
\textbf{Model} & \textbf{Step} & \textbf{Abstract} & \textbf{Animals} & \textbf{Artifacts} & \textbf{Arts} & \textbf{Food \& Bev.} & \textbf{Illustrations} & \textbf{Indoor} & \textbf{Outdoor} & \textbf{Overall} \\
\hline
Baseline & 16 & 24.778 & 28.609 & 27.319 & 28.509 & 26.101 & 27.344 & 27.115 & 26.908 & 27.085 \\
Baseline & 32 & 24.273 & 28.498 & 27.396 & 28.138 & 25.896 & 27.358 & 27.008 & 26.724 & 26.911 \\
Baseline & 48 & 23.687 & 28.301 & 27.225 & 27.835 & 25.661 & 27.400 & 26.800 & 26.436 & 26.668 \\
\hline
Speed-RL & \textbf{16} & \textbf{25.463} & \textbf{28.535} & \textbf{27.142} & \textbf{28.565} & \textbf{25.420} & \textbf{27.319} & \textbf{27.241} & \textbf{26.933} & \textbf{27.077} \\
Speed-RL & 32 & 25.154 & 28.727 & 27.011 & 28.700 & 24.950 & 27.102 & 27.423 & 26.955 & 27.003 \\
Speed-RL & 48 & 24.919 & 28.602 & 26.842 & 28.627 & 24.778 & 27.027 & 27.394 & 26.792 & 26.873 \\
\hline
\end{tabular}}
\caption{PartiPrompts Benchmark HPSv2.0 Scores (Speed-RL vs. Baseline) by Category. Speed-RL achieves a \textbf{+0.409} overall HPSv2.0 reward score with \textbf{3x} fewer steps than the baseline.}
\end{table*}

We evaluate Speed-RL across 16, 32, and 48 steps with the PartiPrompts Benchmark and HPS dataset using HPSv2.0 instead of v2.1. We previously had a minor labeling error and said v2.0 instead of v2.1. Because of this, we provide additional results for v2.0 in Table 4 and 5 (previous results were for v2.1). Overall, Speed-RL beats the Baseline scores for both benchmarks with fewer steps. In the HPS dataset, Speed-RL (16 steps) scores the highest on the Anime style. For the PartiPrompts dataset, Speed-RL (16 steps) scores the highest on the Animals category. 

\section{Derivation of Loss}

Recall that the standard GRPO loss is defined as

\begin{align}
\mathcal{L}_{\text{GRPO-clip}}
&=
\mathbb{E}[
  \frac{1}{G}\sum_{i=1}^G
     \min \left( A_i    r_{i} ,
        A_i\operatorname{clip}\big(r_i, 1-\epsilon, 1+\epsilon\big)\, 
    \right)]
-
\beta \, 
\mathbb{D}_{\mathrm{KL}}\!\left[p_\theta \,\|\, p_{\mathrm{ref}}\right]
,
\\
&\text{where}\quad
r_{i} = \frac{p_{\theta}(x_i \mid c)}{p_{\text{old}}(x_i \mid c)} .
\end{align}

In fully online setup, $p_{\text{old}}=p_\theta$, so the clip is never applied. The loss simplifies to

\begin{align}
\mathcal{L}_{\text{GRPO-noclip}}
&=
\mathbb{E}[
  \frac{1}{G}\sum_{i=1}^G
   A_i    r_{i} 
        ]
-
\beta \, 
\mathbb{D}_{\mathrm{KL}}\!\left[p_\theta \,\|\, p_{\mathrm{ref}}\right]
,
\\
&\text{where}\quad
r_{i}(c,x) = \frac{p_{\theta}(x_i \mid c)}{p_{\text{old}}(x_i \mid c)} .
\end{align}

We plug in our KL Estimator

\begin{align}
\mathbb{D}_{\mathrm{KL}}\!\left[p_\theta \,\|\, p_{\mathrm{ref}}\right]
&=  
\text{H}(p_{\mathrm{ref}}|p_\theta)-H(p_{\mathrm{ref}})\\
&=\mathbb{E}_x[\sum_{i=1}^{L}\sum_{j=1}^{|V|} p_{\mathrm{ref}}(x_i=j|c) \log p_{\theta}(x_i=j|c)-H(p_{\mathrm{ref}})]
\end{align}

where $L=HW$ is the number of image tokens and $|V|$ is the VQ codebook size.  This leads to 

\begin{align}
\mathcal{L}_{\text{GRPO-noclip}}
&=
\mathbb{E}[
  \frac{1}{G}\sum_{i=1}^G
   A_i   \frac{p_{\theta}(x_i \mid c)}{p_{\text{old}}(x_i \mid c)}    
-
\beta \mathcal{L}_{ce}(p_{\mathrm{ref}}(X_i,c),\log p_\theta(X_i,c))]
\\
\label{eq:eq12}
\end{align}

The entropy term, $H(p_{\mathrm{ref}})$ since it does not contribute to gradients. Now consider the gradient 

\begin{align}
    &\nabla_\theta \frac{p_{\theta}(x_i \mid c)}{p_{\text{old}}(x_i \mid c)}= \nabla_\theta \exp (\log p_{\theta}(x_i \mid c) - \log p_{\text{old}}(x_i \mid c))\\
    &=\exp (\log p_{\theta}(x_i \mid c) - \log p_{\text{old}}(x_i \mid c)) \nabla_\theta  (\log p_{\theta}(x_i \mid c)- \log p_{\text{old}}(x_i \mid c))) \\
    &= \exp(0) \nabla_\theta \log p_{\theta}(x_i \mid c) \\
    &= -\nabla_\theta \mathcal{L}_{ce}(\mathbf{x_i},p_\theta(X_i|c))
\end{align}

where $\mathbf{x_i}$ is a one-hot vector. Hence, optimizing \cref{eq:eq12}  is equivalent to optimize  \cref{eq:loss-speed}  in section \cref{sec:stability}.  (Note that the Speed-RL is a minimizing objective, while GRPO is a maximizing objective, so the sign is flipped). Since cross entropy is linear in the first term, we further simplify the loss to the following in our implementation 

\begin{align}
\mathcal{L}_{\text{speed-rl}}
&=
\mathbb{E}[
  \frac{1}{G}\sum_{i=1}^G
 \mathcal{L}_{ce}(A_i \mathbf{x_i}+\beta p_{\mathrm{ref}}(X_i,c),\log p_\theta(X_i,c))]
\\
\label{eq:speed-rl}
\end{align}

Note that technically $A_i \mathbf{x_i}+\beta p_{\mathrm{ref}}(X_i,c)$ are not valid probability vectors as they do not sum to 1 at each token position. Hence, $\mathcal{L}_{ce}$ is a mere algebraic notation rather than a true cross entropy between two probabilities.

\section{Datasets}


We obtain the prompts for RL training from BLIP3o-60k \cite{chen2025blip3} datasets, which contains the prompts from MSCOCO \cite{lin2015microsoftcococommonobjects}, JourneyDB \cite{sun2023journeydbbenchmarkgenerativeimage}, and other hand-crafted prompts.  We visualize their distribution in Figure 8 and Table 6. We observe that the base model generates poor-quality images for extremely short prompts like ``a person", which harms the training stability. Furthermore, we observed that prompts with only emojis and prompts with non-English text also harmed the training stability. Hence, we preprocessed the prompts and removed non-English prompts, prompts with only emojis, and prompts less than 7 words long.

Specifically, the prompts contain the following categories:
\begin{itemize}
    \item \textbf{Randomly Sampled Prompts from JourneyDB}. These are long prompts with detailed descriptions of highly creative and complex scenes, such as characters from movies in different forms and styles and fictional backgrounds combining different elements. For example, "A skydiver from America in the 1970s is captured in a freefall pose, showcasing a mohawk hairstyle reminiscent of 'Weird Al' Yankovic's unique style. The skydiver daringly goes without a helmet, while the backdrop blends the futuristic cityscape of Dubai in 2023 with the contemporary skydiving trends. This realistic photograph, taken using a 35 mm lens on Kodakchrome film with ISO 800, f/4, and a shutter speed of 1/30, offers a breathtaking view in stunning 8k FullHD resolution."
    \item \textbf{Randomly Sampled Prompts from Dalle3}. These are long prompts describing an image in detail. For example, "The image showcases a character with a striking appearance. The character has a mohawk-style hairstyle with a vibrant pink hue. The side of their head is shaved, adorned with a metallic band that has a unique symbol on it. Their face is painted with bold red streaks, and they have piercings on their ears. The character is draped in a shiny, red cloak or garment. The background is a dilapidated, post-apocalyptic setting with rusted metal, broken windows, and a hazy atmosphere."
    \item \textbf{Randomly Sampled Prompts from MSCOCO}. These are captions from the MSCOCO dataset, which consists of real photos. For example, "A man riding a wave on a surf board" and "A woman is slicing a cake with her friends."
    \item \textbf{Randomly Sampled Prompts from GenEval}. These are prompts focused on objects and their properties like count, color, position, and co-occurrence. For example, "a photo of a white hair comb and a brown bear" and "a green zebra and a yellow cabinet."
    \item \textbf{Randomly Sampled Prompts from occupation\_1.txt and occupation\_2.txt}. These are prompts describing a human performing a specific occupation. For example, "An accountant in a crisp white shirt and tie, scanning through a computer screen filled with colorful bar graphs, while sipping a steaming cup of coffee in a modern office."
    \item \textbf{Randomly Sampled Prompts from object\_1.txt and object\_2.txt}. The first set of object prompts include a single object described in around 1 to 3 words, such as "Bar Soap" and "Toaster." The second set of object prompts include a description of a scene that may include more than one object. For example, "A close-up of thermometer and hex bolt on a table" and "A person holding curtain rod and engine piston in their hand."
    \item \textbf{Randomly Sampled Prompts from human\_gestures.txt}. These prompts describe humans performing different kinds of actions. For example, "Driving a car", "Opening a door", "Arriving at school", "Listening to music", and "Passing a basketball."

\end{itemize}
\begin{figure}[h]
    \centering
    \includegraphics[width=0.5\linewidth]{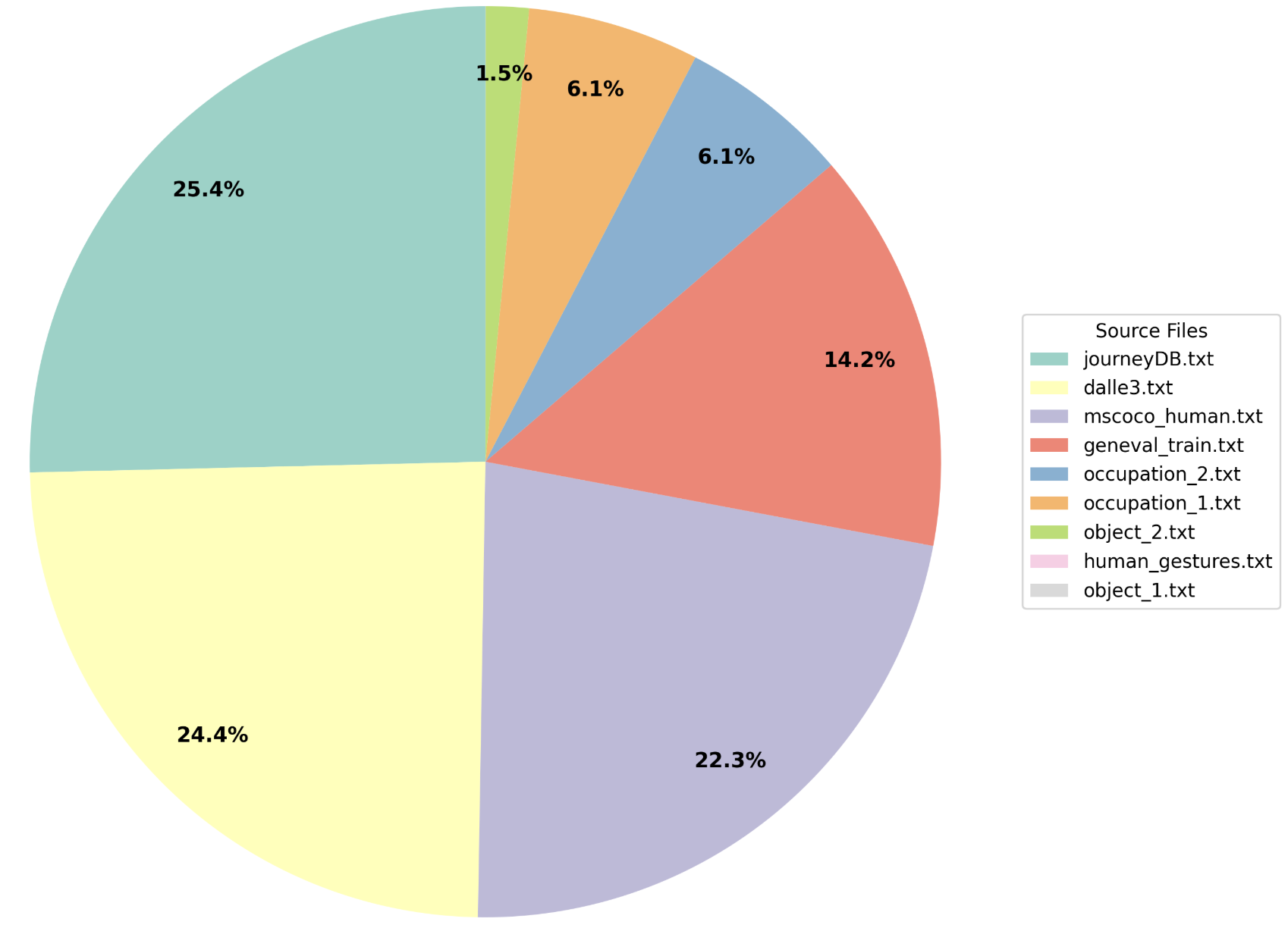}
    \caption{Distribution of selected prompt sources from BLIP3o-60k.}
    \label{fig:blipdist}
\end{figure}

\begin{table}[h]
\centering
\caption{Distribution of prompt sources used from BLIP3o-60k.}
\label{tab:source_distribution}
\begin{tabular}{lr}
\hline
\textbf{Source} & \textbf{Count} \\
\hline
journeyDB.txt & 10,416 \\
dalle3.txt & 10,000 \\
mscoco\_human.txt & 9,154 \\
geneval\_train.txt & 5,848 \\
occupation\_2.txt & 2,500 \\
occupation\_1.txt & 2,500 \\
Decsription of a scene involving several objects (object\_2.txt) & 625 \\
human\_gestures.txt & 2 \\
Single objects described in around 1 to 3 words (object\_1.txt) & 2 \\
\hline
Total & 41,047 \\
\hline
\end{tabular}
\end{table}

\section{Additional Qualitative Examples}
We present additional side-by-side evaluation results in Figure 9. Examples 1 through 9 show a clear improvement in visual quality for Speed-RL (16 steps) compared to the baseline generation with 16 steps. Furthermore, for all nine examples, the 16 step Speed-RL generations are of comparable visual quality to the baseline 48 step generations. In fact, for Examples 7 and 8 Speed-RL's 16 step generations align with the prompt slightly better. In Example 7, Speed-RL's 16 step generation aligns better with key parts of the prompts such as "Maya-style housing" and "Winnie the Pooh." In Example 8, the prompt says "blue sunglasses" which Speed-RL's 16 step generation correctly captures, but the baseline's 48 step generation does not (it generates orange sunglasses instead).

\begin{figure}[h]
    \centering
    \includegraphics[width=0.99\linewidth]{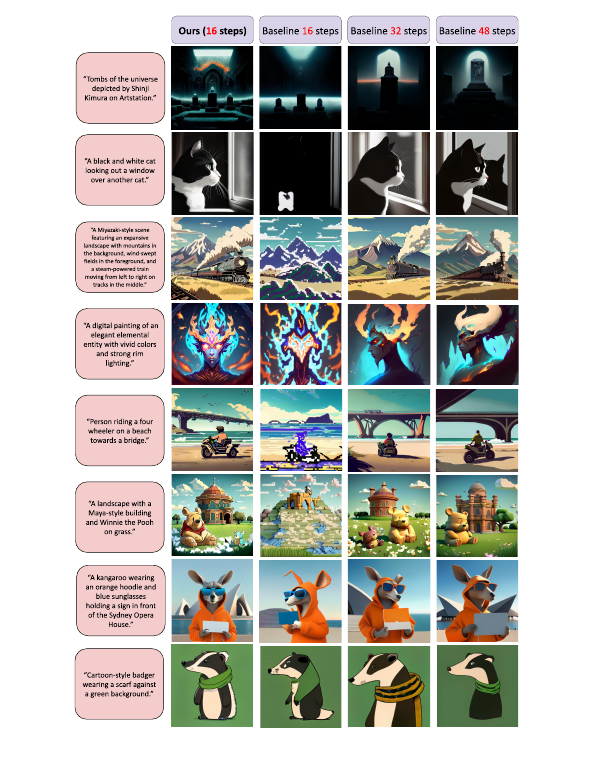}
    \caption{Additional side-by-side examples of Speed-RL with 16 steps and the baseline with 16, 32, and 48 steps.}
    \label{fig:extra-side-by-side}
\end{figure}

\end{document}